\documentclass{article}

\usepackage{PRIMEarxiv}

\usepackage[utf8]{inputenc} 
\usepackage[T1]{fontenc}    
\usepackage{hyperref}       
\usepackage{url}            
\usepackage{booktabs}       
\usepackage{amsfonts}       
\usepackage{nicefrac}       
\usepackage{microtype}      
\usepackage{lipsum}
\usepackage{fancyhdr}       
\usepackage{graphicx}       
\graphicspath{{media/}}     

\usepackage{algorithm}  
\usepackage{algpseudocode}  
\usepackage{amsmath}  

\title{ADLDA: A Method to Reduce the Harm of Data Distribution Shift in Data Augmentation
}

\author{
  Haonan Wang \\
  Weifang University of Science and Technology\\
  \texttt{haonanwang0622@gmail.com} \\
}

\begin{document}
\maketitle

\begin{abstract}
This study introduces a novel data augmentation technique, ADLDA, aimed at mitigating the negative impact of data distribution shifts caused by the data augmentation process in computer vision tasks. ADLDA partitions augmented data into distinct subdomains and incorporates domain labels, combined with domain adaptation techniques, to optimize data representation in the model’s feature space. Experimental results demonstrate that ADLDA significantly enhances model performance across multiple datasets, particularly in neural network architectures with complex feature extraction layers. Furthermore, ADLDA improves the model’s ability to locate and recognize key features, showcasing potential in object recognition and image segmentation tasks. This paper’s contribution provides an effective data augmentation regularization method for the field of computer vision, aiding in the enhancement of robustness and accuracy in deep learning models. 
\end{abstract}

\keywords{Deep Learning \and Computer Vision \and Data Augmentation \and Domain Adaptation}

\section{Introduction}

Deep neural networks have been widely applied in computer vision tasks and have achieved tremendous success. However, due to the general absence of explicit modeling for specific tasks, neural networks are prone to overfitting and exhibit poor robustness. Despite numerous attempts to improve neural network architectures, loss functions, and training algorithms, it has been proven that the most effective way to enhance the robustness of deep neural networks is still to provide a higher quality dataset, which typically means more data \cite{sun2017revisiting, halevy2009unreasonable}. Regrettably, collecting and annotating more data usually implies higher costs and, in many fields such as medical imaging, involves complex social issues, making training with massive data impractical. Therefore, data augmentation techniques are crucial in computer vision tasks, as they generate more useful data from existing datasets, thereby improving the quality of training data and effectively preventing overfitting during neural network training \cite{shorten2019survey}.

Traditionally, we apply methods such as rotation, flipping, color perturbation, and noise injection to training data to generate new data for training. However, these methods may introduce a shift in the data distribution and result in features purely induced by data augmentation, which are irrelevant to the task, participating in the neural network’s task modeling. Inspired by some domain adaptation (DA) works \cite{hashimoto2020multi}, we explored a method to model this phenomenon and reduce its harm during model training, which we call Adversarial Domain Label Data Augmentation (ADLDA). Unlike traditional data augmentation methods, ADLDA divides the data generated by augmentation into different subdomains and assigns domain labels to them.
\begin{figure}[H]
    \centering
    \includegraphics[width=0.75\linewidth]{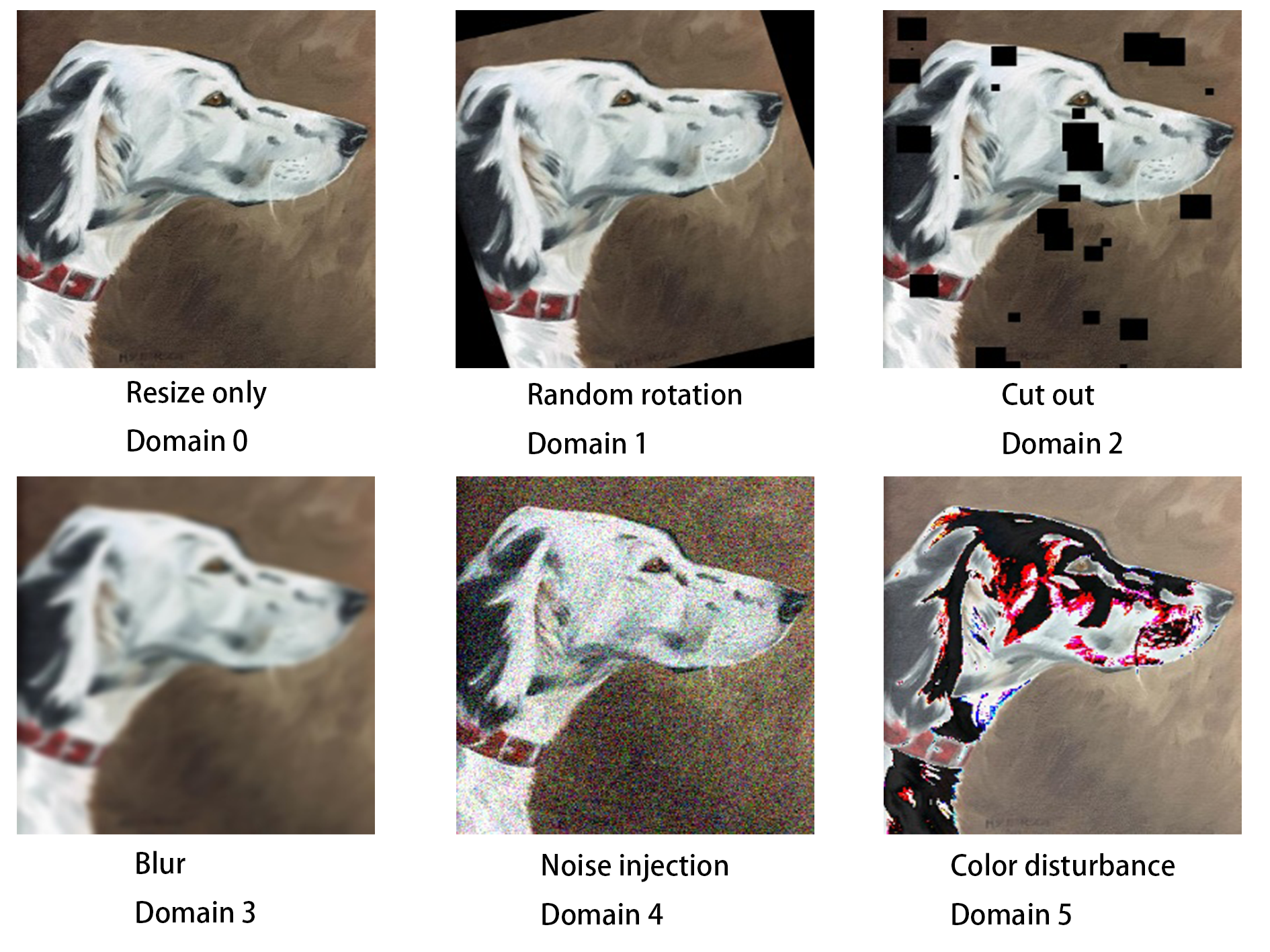} 
    \caption{Using different data augmentation methods and dividing the domains }
    \label{fig1}
\end{figure}

Similar to adversarial domain transfer network structures, a domain discriminator exists after the feature extraction layer in the neural network structure, incorporating domain label loss into the loss function and participating in neural network training. This module can be conveniently added to most neural network structures without affecting the inference speed of the neural network. We validated the performance of several computer vision models on multiple datasets, and the results show that training with the ADLDA module added to the backbone significantly outperforms training with traditional data augmentation methods.

Our main contributions are summarized as follows: We propose a method inspired by adversarial domain adaptation work, ADLDA, to mitigate the harm caused by data distribution shifts during data augmentation. ADLDA can be easily mounted on various models without affecting the model’s inference speed. Experiments show that ADLDA can improve the accuracy of models in classification tasks and enhance the model’s ability to locate and recognize key features.


\section{Related Work}
Attempts to regularize neural networks with data augmentation techniques have a long history. In this study, we employed various data augmentation methods that have been proven effective in the past, such as rotation, distortion, noise injection, or cutout\cite{moreno2018forward, zhong2020random}. We divided the outputs obtained by different data augmentation methods into multiple domains and then used adversarial domain adaptation techniques to reduce the impact of feature distribution differences caused by data augmentation on the model during training.

Domain adaptation is an effective approach aimed at leveraging labeled data from the source domain to improve classification performance in the target domain, where there is a difference in data distribution between the two domains \cite{ganin2015unsupervised}. When there is a shift between the training and testing distributions, the proposed method establishes a mapping between the source domain (training task) and the target domain (testing task), allowing the classifier learned for the source domain to be applied to the target domain.

The main idea of domain adaptation is to bring the data of the source and target domains closer in the feature space through technical means, thereby reducing the impact of domain shift. Recently, domain adaptation methods based on adversarial learning have attracted widespread attention because they can use adversarial loss to align features between the source and target domains while retaining the discriminative nature of the features\cite{jiang2022transferability}. Ben-David et al. have proposed a theory based on optimizing the h-divergence to achieve this purpose in some of their past works\cite{ben2006analysis, ben2010theory}. Y. Ganin et al., inspired by these theories, proposed the DANN structure, which has greatly influenced our work\cite{ganin2016domain}.

Since accurate domain classification is not the ultimate goal in our work, allowing the model to dynamically allocate attention across different domain classes can have a beneficial effect on domain loss. To achieve this, we utilized a multi-head self-attention mechanism\cite{vaswani2017attention}.

\section{Method}

\subsection{Motivation}
For the classification task $T$, with all possible inputs and outputs being $X$ and $Y$, assume there exists an ideal dataset $D\left\{\left(x^{\prime}_i, y_i^{\prime}\right)\right\}_{i=1}^n$, such that the distribution of $D$ perfectly reflects the distribution $X \rightarrow Y$ .  Applying data augmentation functions to $X$ yields an augmented dataset  $A\left\{\left(x^{\prime}_i, y_i^{\prime}\right)\right\}_{i=1}^n$, resulting in a distribution shift between $A$ and $D$. Consequently, $A$ exhibits a distribution shift relative to the distribution $X \rightarrow Y$.
\begin{figure}[H]
    \centering
    \includegraphics[width=0.75\linewidth]{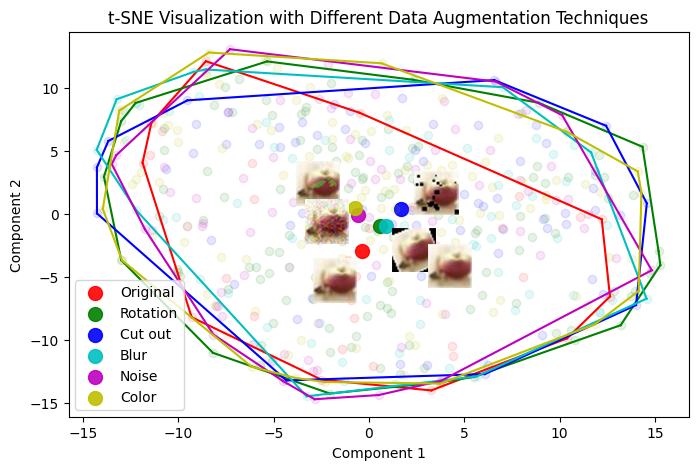}
    \caption{Shift in image features occurs after data augmentation }
    \label{fig2}
\end{figure}

Train the model $f\left(x ; \theta\right)$ on datasets $D$ and $A$ to obtain $f\left(x ; \theta_D\right)$ and $f\left(x ; \theta_A\right)$, respectively. If f can sufficiently learn the distribution of the datasets, then for $x_i, y_i \in X,Y$,
\begin{equation}
\sum_{i=1}^n L\left(f\left(x_i ; \theta_A\right), y_i\right)>\sum_{i=1}^n L\left(f\left(x_i ; \theta_D\right), y_i\right)    
\end{equation}
This implies that for a sufficiently good dataset, data augmentation itself is harmful to model training. However, in practical tasks, the ideal dataset often does not exist, and in most cases, we need to use data augmentation to regularize the training of the model. A viable idea to avoid this harm is to introduce a regularization term $R(\phi(D), \phi(A))$ for the feature extractor $\phi$ during the training of the model, to penalize the representation of the feature space for the shift in data distribution, that is, changing the optimization objective from $L_{\text {task }}(\hat{Y}, Y)$ to
\begin{equation}
    L=L_{\text {task }}(\hat{Y}, Y)+R(\phi(D), \phi(A))
\end{equation}
\subsection{Implementation}

ADLDA can be conveniently applied to various neural networks, and the modification to the network structure will only be limited to adding a domain classifier. In the experiments of this paper, we used an MLP network with multi-head self-attention mechanism involved as the domain classifier, which hardly slows down the training speed of the neural network. During the inference of the neural network, the domain classifier will not participate in the forward propagation, which will not affect the inference speed of the neural network at all.

When using ADLDA, it is necessary to divide the images enhanced by different means into sub-domains during the image enhancement process and add domain labels to them.

Taking the image classification task as an example, the modified neural network has two forward propagation paths. During the forward propagation, the output of the neural network is 
\begin{equation}
\hat{P_y}=G_y\left(G_f\left(x ; \theta_f\right) ; \theta_y\right)
\end{equation}\begin{equation}
\hat{P_d}=G_d\left(A G_f\left(x ; \theta_f\right) ; \theta_d\right)    
\end{equation}
where $\hat{P_y}$ is the predicted class label probability,  $\hat{P_d}$ is the predicted domain label probability, $\theta_f, \theta_y, \theta_d$ are the neural network parameters for the feature extractor, label predictor, and domain classifier respectively, x is the input image, A is the self-attention weight, and $G_f, G_y, G_d$ are the function representations of the feature extractor, label predictor, and domain classifier respectively.

During the inference of the neural network, only the output $\hat{P_y}$ is needed, and there will be no difference in the inference process compared to before the modification. ADLDA mainly improves the performance of the neural network by training the shared parameters $\theta_f$, i.e., the parameters of the feature extractor, in the two forward propagation paths.

In the backpropagation, the optimization target of the neural network is 
\begin{equation}
\left(\widehat{\theta_f}, \widehat{\theta_y}, \widehat{\theta_d}\right) \leftarrow \underset{\theta_f, \theta_y, \theta_d}{\operatorname{argmin}} L_Y\left(\mathbb{Y}_n, \widehat{P_y}\right)-\lambda \sum_{i=0}^n a_i L_D\left(\mathbb{D}_n, \widehat{P_d}\right)     
\end{equation}

\begin{figure}[H]
    \centering
    \includegraphics[width=1\linewidth]{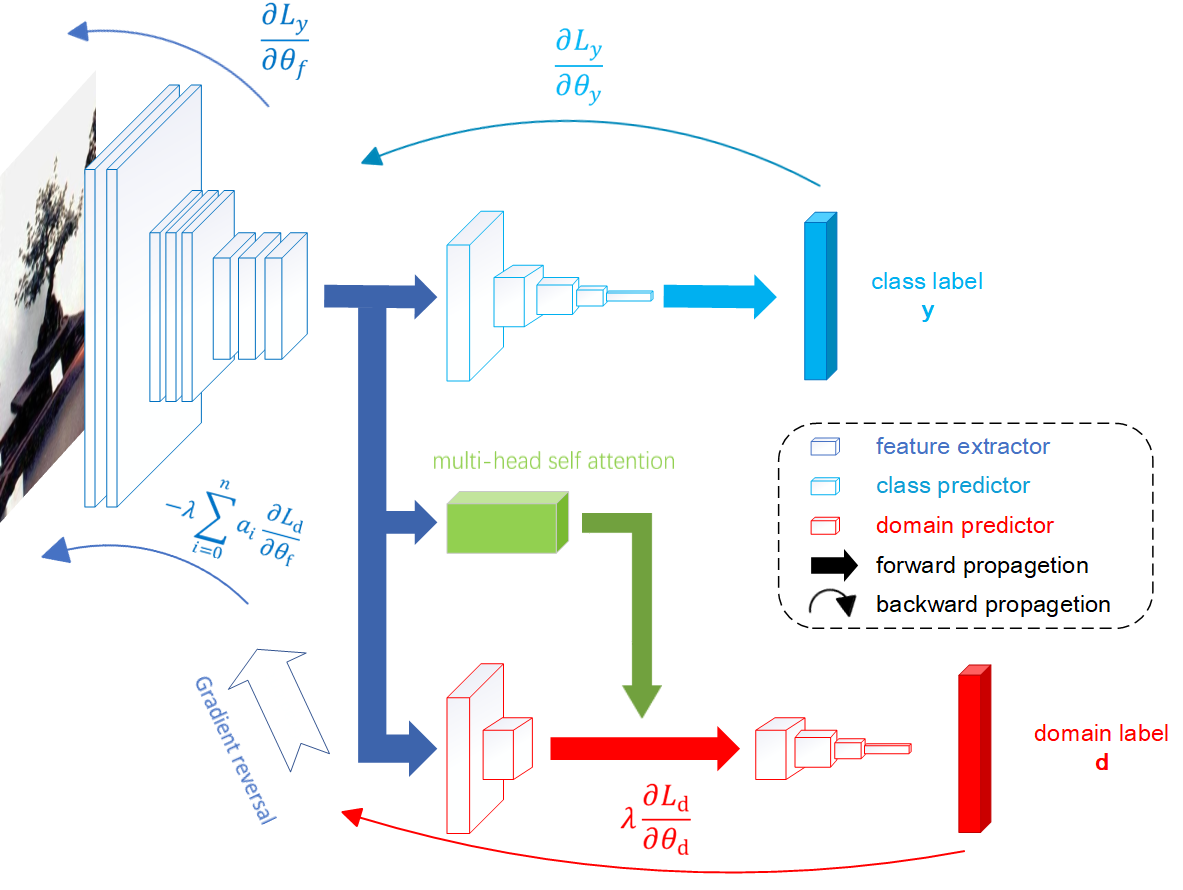} 
    \caption{Neural network architecture }
    \label{fig3}
\end{figure}
In a single batch, the parameters are updated as follows:

 \begin{algorithm}[htb]  
  \caption{ Gradient descent.}  
  \label{alg1}  
  \begin{algorithmic}[1]  
   \State With input image tensor ${x}$, DArate ${\lambda}$, learning rate $\eta$
   \State ${L}_{{Y}} \leftarrow {L}\left(\mathbb{Y}_{{n}}, {G}_{{y}}\left({G}_{{f}}\left({x} ; {\theta}_{{f}}\right) ; {\theta}_{{y}}\right)\right) $
   \State ${L}_{{D}} \leftarrow \left(\mathbb{D}_{{n}}, {G}_{{d}}\left({G}_{{f}}\left({x} ; {\theta}_{{f}}\right) ; {\theta}_{{d}}\right)\right) $
   \State ${L}_{{D}}{ }^{\prime} \leftarrow \sum_{i=0}^n {a}_i {L}_{{D}} $
   \State ${\theta}_{{y}} \leftarrow {\theta}_{{y}}-\mathbf{\eta} \frac{\partial L_Y}{\partial {\theta}_{{y}}} $
   \State ${\theta}_d \leftarrow {\theta}_d-\eta \frac{\partial {L}_{{D}}{ }^{\prime}}{\partial {\theta}_{d}}$
   \State $\theta_f \leftarrow {\theta_f}-\eta\left(\frac{\partial {L_Y}}{\partial {\theta_f}}-\lambda \frac{\partial {L}_{{D}}{ }^{\prime}}{\partial {\theta_f}}\right) $
  \end{algorithmic}  
\end{algorithm}  

\section{Experiment}

\subsection{Dataset}
We validated the impact of the ADLDA module on model accuracy on four commonly used image classification datasets: Caltech256 \cite{griffin2007caltech}, Tiny ImageNet \cite{le2015tiny}, CIFAR-10, and CIFAR-100 \cite{krizhevsky2009learning}. Both CIFAR-10 and CIFAR-100 contain a large number of images with a resolution of 32 x 32 pixels. The CIFAR-10 image classification benchmark includes 50,000 training images and 10,000 test images. The dataset has 10 categories, with 6,000 images per category, of which 5,000 are used for training and 1,000 for testing. CIFAR-100 contains 100 categories, with 600 images per category, of which 500 are used for training and 100 for testing. Each category belongs to one of 20 superclasses. Tiny ImageNet is a subset of ImageNet, containing 200 categories, each with 500 training images, 50 validation images, and 50 test images, with a resolution of 64 x 64 pixels. Since the labels for the test images in Tiny ImageNet are not available, we use the accuracy of the validation images as the experimental result. Compared to running the full ImageNet dataset, Tiny ImageNet requires fewer resources and infrastructure. The Caltech256 dataset contains 30,607 real-world images of varying sizes, covering 257 categories (including 256 object categories and one additional clutter category). Each category has at least 80 images. This dataset is a superset of the Caltech-101 dataset. Since there is no official standard training and test set division provided, we randomly selected 20\% of the images as the test set.
\subsection{Model Accuracy Validation}
We mounted the ADLDA module on five different neural networks with different structures: ViT \cite{dosovitskiy2020image}, EfficientNet \cite{tan2019efficientnet}, DenseNet \cite{huang2017densely}, ResNet \cite{he2016deep}, and VGG \cite{simonyan2014very}, and evaluated the models on multiple datasets. These neural networks are all based on models provided by torchvision, with modifications. We removed the fully connected layers of these models and used the remaining parts as the feature extractor introduced earlier, connecting an MLP as the class predictor and domain predictor. Additionally, based on the characteristics of the datasets, we made certain modifications to the scale of the convolutional kernels of the models without changing the model structure.

 Since the modified class predictor differs from the fully connected layers used in the original models, to prevent this factor from interfering with the experimental results, the baseline does not directly use the models provided by torchvision. Instead, in the modified models, the DA rate, i.e., , is set to 0, so that the domain classification loss does not participate in backpropagation. Regardless of whether the DA rate is set to 0, the image data involved in training will be augmented in the same way to avoid the effect of data augmentation itself on the experimental results.

 Since the ViT model is very difficult to converge on small-scale datasets, we used some of its weights pre-trained on ImageNet1k for transfer learning. Therefore, we did not evaluate the accuracy of this model on the Tiny ImageNet dataset. For other models, no pre-trained weights were used. Since we made certain modifications to the models, we could not directly use the hyperparameters used in other papers. We could only ensure that the same hyperparameters were used for each control group. 
 

\begin{table}[H]
\centering
\caption{Validation accuracy of multiple models on multiple datasets, all test accuracies are the average of more than 3 training times, FT means that the pre-trained model was used for Fine-Tune \label{tab1}}
\begin{tabular}{@{}cccccc@{}}
\toprule
DataSet & Model & Input Tensor Shape & Traing images/Test images & Baseline   & ours       \\ \midrule
CIFAR10       & VGG11& 3×32×32& 50k/10k& 89.91& 90.16\\
CIFAR10       & ResNet18& 3×32×32& 50k/10k& 93.16& 94.22\\
CIFAR10       & DenseNet121& 3×32×32& 50k/10k& 91.03& 91.52\\
CIFAR10       & ViTb32(FT)& 3×224×224& 50k/10k(FT)& 95.72& \textbf{96.81}\\
CIFAR100       & VGG11& 3×32×32& 50k/10k& 59.06& 58.84\\
CIFAR100       & ResNet18& 3×32×32& 50k/10k& 68.22& \textbf{69.68}\\
CIFAR100       & DenseNet121& 3×32×32& 50k/10k& 65.75& 66.54\\
Tiny-ImageNet       & ResNet34& 3×64×64& 100k/10k& \textbf{60.82}& 60.43\\
Tiny-ImageNet       & EfficientNetB0& 3×64×64& 100k/10k& 50.40& 51.32\\
Caltech256       & ResNet50& 3×224×224& 24053/6132& 40.67& 41.58\\
Caltech256       & EfficientNetB3& 3×224×224& 24053/6132& 38.70& 38.96\\
Caltech256       & EfficientNetB3(FT)& 3×224×224& 24053/6132(FT)& 87.02& \textbf{88.64}\\
Caltech256       & ViTb32(FT)& 3×224×224& 24053/6132(FT)& 82.34& 83.38\\ \bottomrule
\end{tabular}
\end{table}

We have noticed that on multiple datasets, the ADLDA method has a more significant effect on ViT models with complex feature extraction layers. However, on models like the VGG network, which have relatively simple feature extraction layers, the performance improvement brought by ADLDA is quite limited. This phenomenon may be due to the principle of the ADLDA method, which regularizes the feature extraction layer through a domain classification loss function. Therefore, when the feature extraction layer itself is simple, the performance improvement space brought by regularization is limited. In contrast, for network structures like ViT, which have more complex feature extraction layers, the ADLDA method can more effectively promote the model to learn richer and more abstract feature representations, thereby improving classification accuracy. 

\subsection{Improvements in Other Aspects of the Model}
By plotting the GradCAM images of the ResNet50 model during the inference process, we can intuitively see which regions are assigned higher importance when the model makes predictions. It is observed that after applying the ADLDA method, there is a significant improvement in the GradCAM heatmaps of the ResNet50 model. Specifically, the heatmaps more accurately cover the target objects in the images that need to be classified. This phenomenon indicates that the ADLDA method has enhanced the model’s ability to locate and recognize key features, demonstrating potential in the fields of object recognition and image segmentation.  

\begin{figure}[H]
    \centering
    \includegraphics[width=0.75\linewidth]{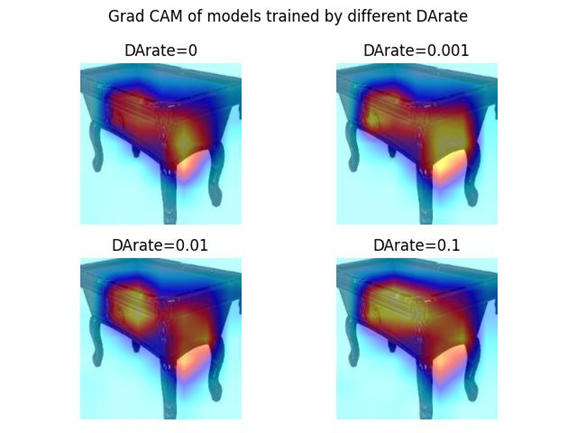} 
    \caption{CAM of the trained models under different DARates}
    \label{fig4}
\end{figure}

\section{Conclusion}

We proposed a new data augmentation method in our study, which divides the output into different domains according to various augmentation techniques during the data augmentation process. During model training, a domain classification loss is used to regularize the feature extraction layer, thereby mitigating the impact of distribution shifts that occur during data augmentation on model training. 

 We trained and evaluated various neural network architectures with different structures on multiple datasets. The experimental results show that the ADLDA module can effectively regularize the feature extraction layer of the model by using the domain classification loss function, thereby significantly improving the classification accuracy of various types of neural networks on multiple datasets. This method is particularly effective for neural networks with complex feature extraction layers, such as ViT. 

 From the effects of GradCAM, we can observe that the ADLDA module effectively enhances the model’s ability to locate and recognize key features. This implies that this method has potential in tasks such as object recognition or image segmentation. In future work, we will continue to explore the effects of this method in other computer vision tasks.
\bibliographystyle{unsrt}  
\bibliography{references}

\end{document}